\documentclass[10pt,twocolumn,letterpaper]{article}

\usepackage{3dv}
\usepackage{times}
\usepackage{epsfig}
\usepackage{graphicx}
\usepackage{amsmath}
\usepackage{amssymb}
\usepackage{tabularx}
\usepackage{array}
\usepackage{subcaption}
\usepackage[export]{adjustbox}[2011/08/13]
\usepackage{float}

\threedvfinalcopy 

\def\threedvPaperID{292} 

\ifthreedvfinal\fi
\begin{document}

\title{Temporal LiDAR Frame Prediction for Autonomous Driving}

\author{David Deng \qquad Avideh Zakhor\\
UC Berkeley\\
{\tt\small \{davezdeng8, avz\}@berkeley.edu}
}

\maketitle

\begin{abstract}
Anticipating the future in a dynamic scene is critical for many fields such as autonomous driving and robotics. In this paper we propose a class of novel neural network architectures to predict future LiDAR frames given previous ones. Since the ground truth in this application is simply the next frame in the sequence, we can train our models in a self-supervised fashion. Our proposed architectures are based on FlowNet3D and Dynamic Graph CNN. We use Chamfer Distance (CD) and Earth Mover's Distance (EMD) as loss functions and evaluation metrics. We train and evaluate our models using the newly released nuScenes dataset, and characterize their performance and complexity with several baselines. Compared to directly using FlowNet3D, our proposed architectures achieve CD and EMD nearly an order of magnitude lower. In addition, we show that our predictions generate reasonable scene flow approximations without using any labelled supervision. \\
   
\end{abstract}

\section{Introduction}

Several autonomous driving companies such as nuTonomy, Waymo, and Lyft have recently released large scale datasets \cite{nuscenes2019, waymo} with high resolution LiDAR point clouds and numerous annotations. Perhaps more interestingly, they provide high frequency temporal LiDAR frames, or LiDAR ``videos", opening doors to the use of temporal information in LiDAR point cloud analysis. This paper aims to take advantage of the temporal data offered by the newly released datasets and tackles the point cloud prediction task using deep learning methods, that is, given a sequence of past point cloud frames, to predict future frames. While this task can easily be applied to other contexts, we implement and evaluate our models specifically in the context of autonomous driving and LiDAR. Much of the autonomous driving problem has to do with helping cars anticipate events and avoid collisions. To this end, predicted point clouds could be used to enhance object tracking pipelines, or to generate preliminary region proposals for future detections. Furthermore, no data labelling is required for this task; the label is the next frame of the LiDAR sequence, so our networks can be trained in a self-supervised fashion. 

Extracting temporal information from point clouds is a difficult task and little work has been done in this area. Conventional architectures for sequential data such as LSTMs cannot be applied directly to point clouds due to their irregular structure. Voxelizing the point cloud could work around this, but it reduces the resolution of the point cloud and introduces quantization artifacts into the prediction. In this paper, we explore a number of end to end architectures that operate directly on point clouds to extract temporal features and predict future frames. In particular, we design a general architecture framework and implement and evaluate several of its variations. Through experiments we evaluate the performance and complexity of our models and assess the tradeoffs of various architectural design choices. 

\begin{figure*}
\begin{center}
\includegraphics[width=\textwidth]{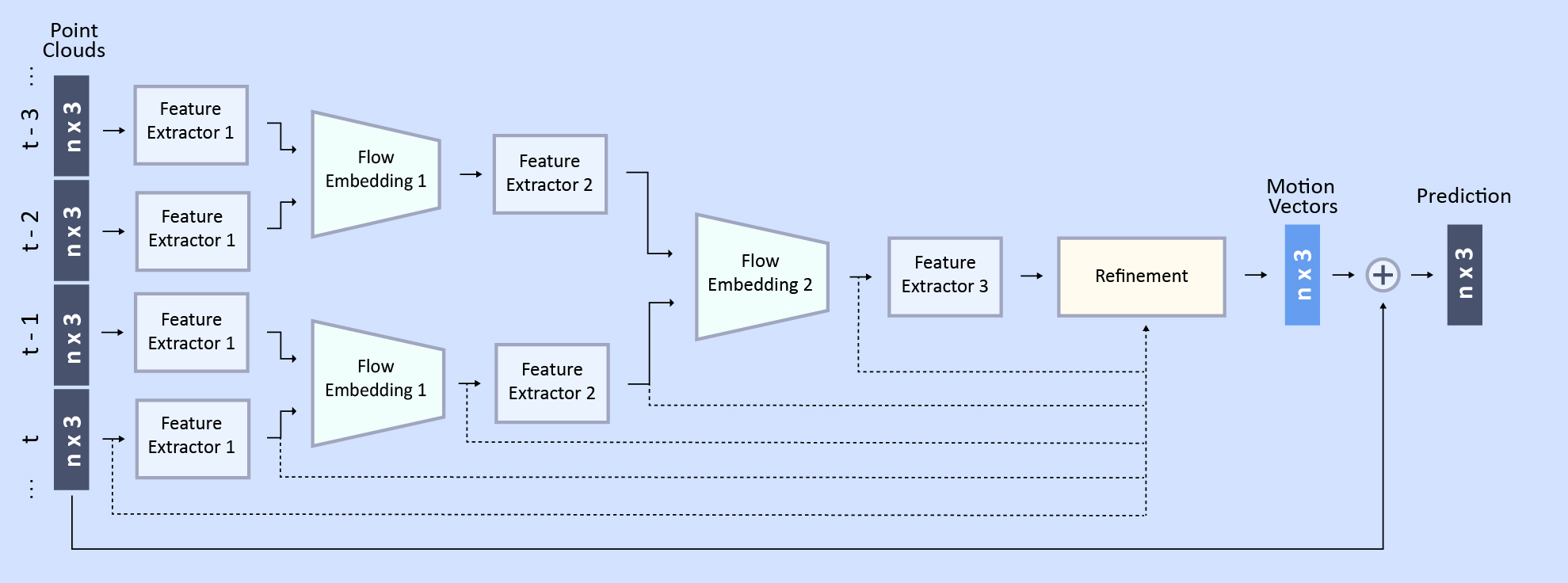}
\vspace{-20pt}
\end{center}
   \caption{\textbf{Generic architecture framework.} Our framework takes in the past 4 frames and generates motion vectors to predict the next frame. The specific architecture is determined by which feature extractor is used and whether or not downsampling is used. The refinement module may use any of the previous learned features, as indicated by the dashed lines. For specific architecture details refer to Section 3 and the supplementary material.}
\label{fig:architecture}
\vspace{-5pt}
\end{figure*}


\section{Related Work}

\subsection{Deep Learning on Point Sets}

\noindent \textbf{Feature Extraction} \hspace{2pt}
Early attempts at learning on point clouds involved converting them to voxels and using CNNs to extract features. Works such as \cite{voxnet, voxelnet} are able to achieve reasonable results on classification and detection. However, these methods tend to be limited by their high computation costs and quantization artifacts. In 2017, Qi et al. proposed PointNet \cite{pointnet}, the first deep learning architecture that operated directly on point clouds by utilizing symmetric functions such as max pooling to make the network invariant to input permutations. Later works \cite{pointnet++, kpconv, pointcnn} would take advantage of the local structures in point clouds to learn local feature descriptors. Dynamic Graph CNN (DGCNN) \cite{dgcnn} is a special case among these. While most architectures group points and learn features based on Euclidean distance, DGCNN dynamically groups them based on their distance in the feature space, allowing greater expressive power and more useful local features. We take advantage of these properties in our proposed architectures. 

\medskip
\noindent \textbf{Generative Models} \hspace{2pt}
Recently, generative models such as GANs and autoencoders have been applied to point clouds. This introduces the need for a point cloud similarity metric. \cite{pointsetgeneration} proposes two such metrics, i.e.\ Chamfer Distance (CD) and Earth Mover's Distance (EMD). We train our networks to minimize these distance functions over our predicted point clouds and the ground truth. 

\subsection{Temporal Learning on Point Sets}

\noindent \textbf{Temporal Learning with Neural Networks} \hspace{2pt}
Recurrent networks are well studied and known for their application on sequential data. In the past decade they have been applied to the task of video prediction \cite{prednet}, which is conceptually similar to our task. However, recently feedforward networks have been shown to perform just as well, if not better than recurrent networks on sequential data \cite{bai2018empirical}. They do not suffer from vanishing gradients and other training instabilities common in recurrent networks, tend to be faster and more parallelizable, and typically require less memory during training. Similar to recurrent networks, feedforward models have also been applied to temporal video processing and prediction tasks \cite{phd}. Inspired by this, we choose to use feedforward networks for our proposed architectures. 

\medskip
\noindent \textbf{Temporal Learning on Point Sets} \hspace{2pt}
In 2018, Liu et al. published Flownet3D \cite{flownet3d}, which takes two point clouds and predicts the scene flow from one to another. Our proposed architectures are inspired by FlowNet3D but are tailored more specifically to the prediction task and leverage more modern feature extractors. In the short time following FlowNet3D, a handful of other works tackled the point cloud scene flow problem as well \cite{HPLFlowNet, meteornet}, while others utilized temporal information in object detection \cite{yolo4d}. Recently, a few papers have tackled point cloud prediction. \cite{pointrnn} introduces a recurrent architecture which operates directly on point clouds, while \cite{100000pts} extracts global feature vectors from the point clouds, uses an LSTM to learn temporal patterns, and decodes the LSTM output into the predicted point cloud.

\section{Method}

Our work aims to predict future point cloud frames given previous ones. In this section we introduce our architecture framework, describe several variants of this framework, and share our training details. 

\subsection{Architecture Framework}

Our generic architecture framework is illustrated in Figure \ref{fig:architecture}. Given the past 4 frames of point clouds $x_{t-3}, ..., x_t$, our network generates $x^*_{t+1}$, the prediction for $x_{t+1}$. While 2 frames is already sufficient for recovering the velocity of objects in a scene, we condition on 4 frames because at least 3 is required to recover second order dynamics i.e.\ acceleration, and additional frames provide contextual information. At each stage of the network, we first extract pointwise features from every frame and then learn the dynamics of the scene using the flow embedding layer introduced in \cite{flownet3d}. After two such stages, we extract features on the single remaining point cloud and push the features through a refinement module. The refinement layers output a motion vector for each point in $x_t$. We add these predicted vectors to $x_t$ to generate $x^*_{t+1}$. We chose to reformulate our task into a motion prediction problem rather than directly regressing the output point cloud because this adds more interpretability to the output of our model, and we found it to be easier to optimize as well. Point clouds in the more distant future can be predicted by applying the model autoregressively ie. by feeding $x^*_{t+1}$ back into the model as a pseudo groundtruth for $x_{t+1}$, using this to predict $x^*_{t+2}$, and so on.

We explore two approaches of modulating the framework to create different architectures. The first approach is the choice of feature extractor. In this paper we experiment with the PointNet++ layer \cite{pointnet++} and the EdgeConv layer from \cite{dgcnn}. The second approach is whether or not to downsample the features throughout the initial stages of the network and upsample back to the original resolution in the refinement stage. We discuss these in further detail in the following sections. 

\subsection{Feature Extractors}

\noindent \textbf{PointNet++ Layer} \hspace{2pt}
The PointNet++ layer takes an input point cloud $\{p_1, p_2, ..., p_n\}, p_i \in R^3$ and its features $\{f_1, f_2, ..., f_n\}, f_i \in R^c$, and outputs a new set of features $\{f'_1, f'_2, ..., f'_{n'}\}, f'_i \in R^{c'}$. For each point, it groups its neighbors within a given radius and applies the PointNet operation to that local region, producing a new feature vector. More specifically: 
\begin{equation}
f'_i = \max_{j\mid\: ||p_j-p_i|| \leq r} h_\theta(f_j, p_j-p_i)
\label{equation: pn++}
\end{equation}
where $r$ is the radius of the ball query, $h_\theta$ is a multilayer perceptron (MLP) with weights $\theta$ and input and output dimensions $R^{c+3}$ and $R^{c'}$ respectively, and $\max$ is the element-wise maximum function

\medskip
\noindent \textbf{EdgeConv Layer} \hspace{2pt}
The EdgeConv layer takes in only the point features $\{f_1, f_2, ..., f_n\}, f_i \in R^c$, and outputs a new set of features $\{f'_1, f'_2, ..., f'_{n'}\}, f'_i \in R^{c'}$. For each point, it finds its k nearest neighbors (KNN) in the \emph{feature space} and applies a MLP across the point's original feature and the difference between it and its neighbor's features. It then groups all k feature vectors with max pooling. More specifically: 
\begin{equation}
f'_i = \max_{j = 1...k} h_\theta(f^j_i-f_i, f_i)
\end{equation}
where $f^j_i$ is the feature of the jth nearest neighbor of $f_i$ in the feature space, $h_\theta$ is a MLP with input and output dimensions $R^{2c}$ and $R^{c'}$ respectively, and all other symbols are defined as in Eq. \ref{equation: pn++}. 

\medskip
These two feature extractors are computationally quite similar; they compute local features and achieve permutation invariance using the symmetric max pooling function. The main difference lies in how they define locality. The PointNet++ layer groups points in the original Euclidean space whereas the EdgeConv layer groups points dynamically in the computed feature space. This allows the EdgeConv layers to have an effectively larger receptive field and compute potentially more descriptive local features, giving it the upper hand in terms of point cloud classification and segmentation performance. 

\begin{table}[t]
\begin{center}
\begin{tabular}{|p{1.9cm}||p{2.9cm}|p{2.2cm}|}
\hline
\textbf{Module} & \textbf{Downsampling} & \textbf{No \newline Downsampling}\\
\hline
\hline
Feature \newline Extraction 1 & SR=0.25$\times$, \newline mlp=[128, 128] & mlp=[32, 32]\\
\hline
Flow  \newline Embedding 1& SR=1$\times$, \newline mlp=[128] & 
mlp=[32] \\
\hline
Feature \newline Extraction 2& SR=0.25$\times$, \newline mlp=[256, 256] &  mlp=[64, 64]\\
\hline
Flow  \newline Embedding 2& SR=1$\times$, \newline mlp=[256] & 
mlp=[64]\\
\hline
Feature \newline Extraction 3& SR=0.2$\times$, \newline mlp=[512] &  mlp=[128] \\
\hline
Refinement & {Upconv1}: SR=5$\times$, mlp1=[512], mlp2=[512] &
    {mlp}: [512, 256, 128, 3]\\
    & {Upconv2}: SR=4$\times$, mlp1=[512], mlp2=[512] & \\
    & {FeatProp}: SR=4$\times$, mlp=[256] & \\
    & {mlp}: [256, 128, 3] & \\

\hline
\end{tabular}
\end{center}
\vspace{-10pt}
\caption{\textbf{Downsampling vs. No Downsampling.} Architectural comparison between downsampling and non downsampling models. SR = sampling rate, Upconv refers to the Set Upconv module introduced in \protect\cite{flownet3d}, and FeatProp refers to the feature propagation module from \protect\cite{pointnet++}. For additional architecture details, refer to the supplementary.}
\label{table:downsample}
\vspace{-10pt}
\end{table}

\subsection{Downsampling}
In the original papers, the EdgeConv layer maintains the size of the point cloud while the PointNet++ layer downsamples point clouds by sampling a subset of the points with iterative furthest point sampling (FPS) and computing features for these points alone. Since our model predicts motion vectors for each point, downsampled features need to be upsampled back to size of the original point cloud. \cite{flownet3d} accomplishes this using the Set Upconv layer. Given a lower resolution point cloud $\{p_1, p_2, ..., p_{n'}\}, p_i \in R^3$ and its features $\{f'_1, f'_2, ..., f'_{n'}\}, f_i \in R^c$, as well as a previously computed higher resolution point cloud $\{p_1, p_2, ..., p_n\}, p_i \in R^3$ with its features $\{f_1, f_2, ..., f_n\}, f_i \in R^c$, the Set Upconv layer applies the PointNet++ operation to each point in the higher resolution point cloud by grouping the points from the lower resolution point cloud. These features are then further processed with an additional MLP. More precisely: 
\begin{equation}
f^*_i = h_{\theta_2}(\max_{\substack{j\mid\: ||p_j-p_i|| \leq r,\\ p_j \in p'}} h_{\theta_1}(f_j,p_j-p_i), f_i)
\end{equation}
where $p'$ indicates the set of points in the lower resolution point cloud. At the final upsampling layer, \cite{flownet3d} uses the Feature Propagation layer from \cite{pointnet++}, which replaces the first MLP with an inverse distance weighted average interpolation. We also adopt this upsampling strategy in our PointNet++ based architectures. 

Although EdgeConv has not been used with downsampling, we also investigate this architecture configuration for point cloud prediction and develop novel downsampling and upsampling modules. To do this, we draw parallels between EdgeConv and the PointNet++ layer and design the sampling scheme in the spirit of these parallels. To downsample the points, rather than using FPS in Euclidean space, we compute it in the feature space. To upsample, we utilize the Set Upconv layer, but similarly, we group the points in the previously computed feature space rather than Euclidean space. 

Downsampling is beneficial because it helps reduce the computational complexity of the network. However it also reduces the resolution of the features and creates ambiguity when upsampling. Thus networks that use downsampling need to be larger in order to resolve these ambiguities and have sufficient feature content at the architecture bottleneck. For a comparison between downsampling and non downsampling architectures, please refer to Table \ref{table:downsample}.

\begin{table}[t]
\begin{center}
\begin{tabular}{r|p{2.4cm}|p{2.7cm}|}
 & Downsampling & No Downsampling\\
\hline
PointNet++ & FlowNet3D \cite{flownet3d}, \newline PN++ w/ DS & PN++ w/o DS \\
\hline
EdgeConv & EC w/ DS & EC w/o DS\\
\hline
\end{tabular}
\end{center}
\vspace{-10pt}
\caption{\textbf{Architecture classification.} Primary architectural differences between our proposed architectures and FlowNet3D.}
\label{table:differences}
\vspace{-10pt}
\end{table}

\subsection{Proposed Architectures}
Modulating the feature extractor and sampling strategy of our framework results in four different architectures as seen in Table \ref{table:differences}: PointNet++ with downsampling (PN++ w/ DS), PointNet++ without downsampling (PN++ w/o DS), EdgeConv with downsampling (EC w/ DS) and EdgeConv without downsampling (EC w/o DS). The exact architecture details are described in the supplementary material. 

\subsection{Loss Functions}

For our loss, we need a function that measures the similarity between two point clouds $P$ and $Q$. Following \cite{pointsetgeneration}, we use CD and EMD: 

\begin{equation}
L_{CD}(P,Q) = \frac{1}{2}(\sum_{p \in P} \min_{q \in Q}||q-p||^2_2+\sum_{q \in Q} \min_{p \in P}||p-q||^2_2)
\end{equation}

\begin{equation}
L_{EMD}(P,Q) = \min_{\phi \in P \rightarrow Q} \sum_{p \in P} ||p-\phi(p)||_2
\end{equation}
where $\phi$ indicates a bijection. 

As in \cite{pointsetgeneration}, we utilize a parallelizable approximation of the true EMD \cite{approxemd}. \cite{generativemodelsforpc} notes that CD does not always remain true to finer, visual similarities between point clouds, and that when optimized, it may overpopulate regions where points are more likely to appear in the ground truth. EMD more accurately captures visual similarity; however, CD still works well at capturing coarser, structural similarities, and we find that is an easier function to optimize. In our loss function, we use a combination of both:

\begin{equation}
L(P, Q) = \alpha L_{CD}(P, Q) + \beta L_{EMD}(P, Q)
\end{equation}
where $\alpha$ and $\beta$ are parameters chosen with cross validation.

\begin{table*}[t]
\begin{center}
\begin{tabular}{|c|c|c|c|c|c|}
\hline
Model & CD ($m^2$) & EMD ($m$) & Model Size (MB) & Runtime (s) & Max Memory Allocated (MB)\\
\hline
\hline
Identity & .2472 & 34.88 & - & - & -\\
\hline
FN3DOOB \cite{flownet3d} & 1.2084 & 91.68 & 14.9 & .2946 & \textbf{669.6}\\
\hline
FN3DA & .1399 & 33.94 & 14.9 & .2946 & \textbf{669.6}\\
\hline
PN++ w/ DS & \textbf{.1381} & \textbf{33.14} & 22.1 & .2635 & 783.5\\
\hline
PN++ w/o DS & .1813 & 37.63 & \textbf{3.7} & .5079 & 1007.2\\
\hline
EC w/ DS & .1837 & 35.49 & 10.2 & \textbf{.2591} & 907.9\\
\hline
EC w/o DS & .1450 & 34.23 & 3.9 & .6198 & 721.3\\
\hline
\end{tabular}
\end{center}
\vspace{-10pt}
\caption{\textbf{Accuracy and complexity of methods.} The table shows the average CD and EMD across the first 5 future frames, as well as the size, runtime, and memory usage of the models.}
\label{table:performance}
\end{table*}

\begin{figure*}[t]
\begin{center}
\vspace{-10pt}
\includegraphics[width=\textwidth]{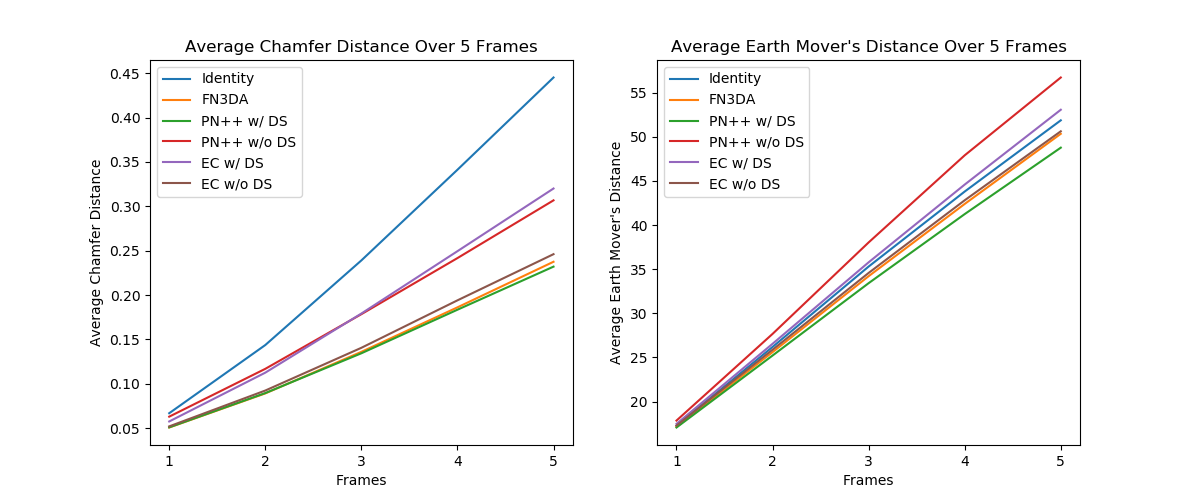}
\vspace{-25pt}
\end{center}
   \caption{\textbf{Average Chamfer Distance and Earth Mover's Distance of our method and baselines.} Plotted over 5 frames. FN3DOOB was omitted to make the other methods more distinguishable.}
\label{fig:results}
\vspace{-5pt}
\end{figure*}

\subsection{Training Details}
There exist two popular approaches for training generative, temporal networks: curriculum approaches \cite{curriculum} and teacher-forcing \cite{teacher_forcing}. In teacher-forcing, the model is trained using only ground truth inputs. However, this prevents the model from learning how to use its own predictions as inputs as it does during test time. Instead, we use a curriculum based approach by training the model with its own predictions as inputs. We first train the model to predict $x^*_{t+1}$, and once that converges, we train on $x^*_{t+2}$, feeding in our predicted $x^*_{t+1}$ back into the model. This way, we slowly increase the difficulty as the model becomes capable of learning harder tasks. We repeat this process until the validation loss no longer decreases upon training the next time step.

We train our models on the nuScenes dataset \cite{nuscenes2019}, a recently released large scale autonomous driving dataset. It contains over 320,000 point clouds from rotating LiDAR scans captured at 20 Hz with over 34,000 points each. However, many of these points are detecting the roof of the ego vehicle, which is of little interest. In addition, the outer most points are extremely sparse and less relevant to potential downstream driving decisions. So we preprocess the point clouds by selecting an annular region of points between the 12,000th and 34,000th point from the origin. This adequately filters out the ego vehicle and fringes of the point cloud. We keep the point clouds in the original coordinate frame from the LiDAR scanner instead of transforming them to the static global frame, choosing only to operate with the raw sensor data. However, an interesting avenue of future work could investigate how utilizing this transformation affects prediction accuracy.

Our models use leaky ReLU activations with slope 0.2 followed by batch normalization, except for the layer directly preceding the output. To train them we use the AdamW optimizer \cite{adamw} with decoupled weight decay and $L_2$ regularization and employ a cosine annealing learning rate scheduler with restarts each time we advance to the next time step \cite{cosineannealing}. We chose values of 1 and 0.02 for $\alpha$ and $\beta$ in the loss function, respectively. For $t+1$, we use a max learning rate of .001 and find that the models converge in 2 epochs, and for all other time steps we use a max learning rate of .0001 and find they typically converge after 1 epoch. We train until $t+3$, after which the loss no longer decreases. 

One of the challenges of working with large point clouds is the computational cost. For $t+1$ we used a batch size of 4; however, training future time steps linearly increases the memory usage, dropping the batch size and making our batch normalization layers ineffective. To address this, we trained with regular batch normalization for $t+1$, but for future time steps we instead normalize our features using the learned, running estimates of the mean and variance from the first time step. When training on time steps beyond $t+1$, the weights of the network do not change as much as they do during the initial step. Thus, the learned statistics from $t+1$ are still adequate estimates beyond $t+1$, allowing us to train with smaller batch sizes while preserving accuracy. 

\begin{figure*}[hbtp]
\begin{center}
\vspace{-3pt}
\includegraphics[width=\textwidth, center]{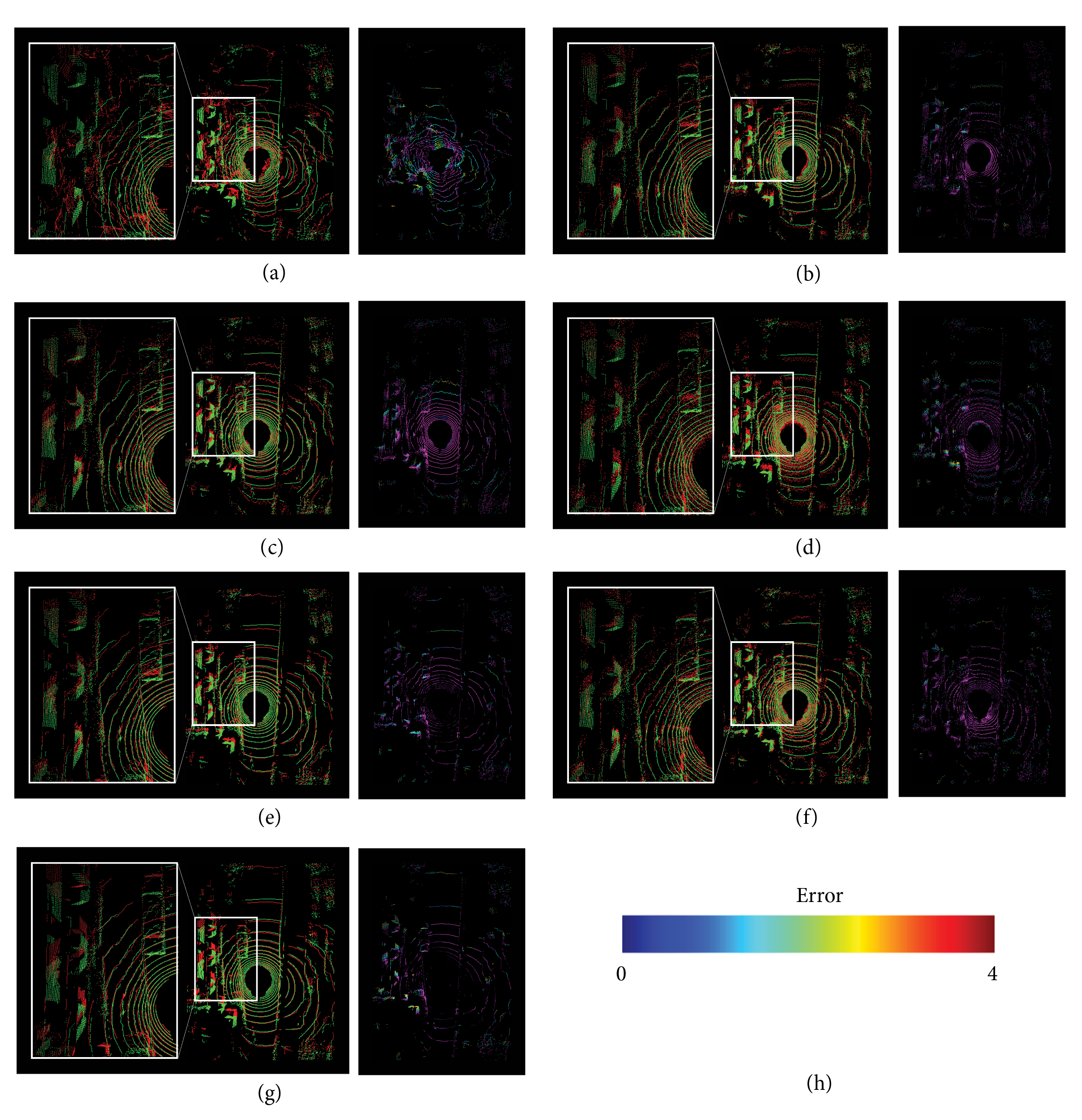}
\vspace{-35pt}
\end{center}

    \caption{\textbf{Visualizations.} Error visualizations for (a) FN3DOOB, (b) FN3DA, (c) PN++ w/ DS, (d) PN++ w/o DS, (e) EC w/ DS, (f) EC w/o DS, (g) Identity on $t+5$. For each method, the middle picture shows the ground truth in green and the prediction in red; the left picture zooms in on a region of interest in the middle picture; the right picture shows the squared distance between each point in the prediction and its nearest neighbor in the ground truth point cloud, with the color bar in (h) indicating the scale of the error. Refer to the identity visualization for scene context.}
\label{fig:visualization}
\vspace{-5pt}
\end{figure*}

\begin{figure*}[hbtp]
\begin{center}
    \includegraphics[width=\textwidth, center]{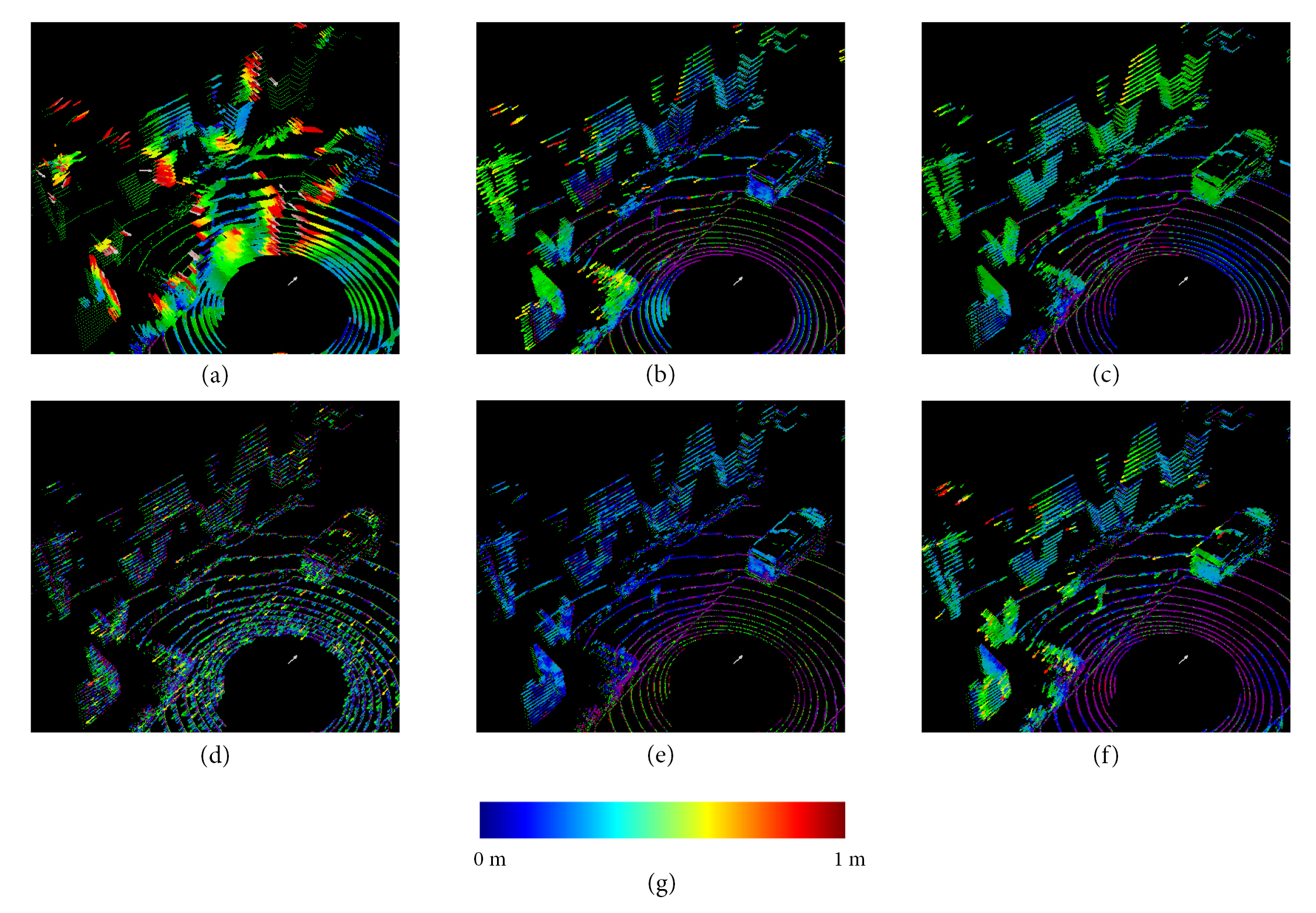}
\vspace{-30pt}
\end{center}
    \caption{\textbf{Flow visualization.} Predicted motion vectors for (a) FN3DOOB, (b) PN++ w/ DS, (c) EC w/ DS, (d) FN3DA, (e) PN++ w/o DS, (f) EC w/o DS. The arrows in the visualization indicate our model's predicted motion vectors, and the color corresponds to the magnitude, as indicated by the color bar in (g). Vectors beyond the range of the color bar are omitted. }
\label{fig:flow_vis}
\vspace{-5pt}
\end{figure*}

\section{Experiments}

In this section we evaluate the performance of our models against several competitive baselines. We provide quantitative analysis on the accuracy and complexity of our model, as well as qualitative visualizations. 

\subsection{Baselines}

\noindent \textbf{Identity} \hspace{2pt}
Our first, and most naive baseline, is to use $x_t$ as our prediction for $x_{t+1}$. We call this the identity baseline. 

\medskip
\noindent \textbf{FlowNet3D Out of the Box (FN3DOOB)} \hspace{2pt}
This baseline takes FlowNet3D from the original paper trained on FlyingThings3D \cite{flyingthings3d}, computes the scene flow from $x_t$ to $x_{t-1}$, and subtracts that from $x_t$ to generate a prediction for $x_{t+1}$. Note that predicting the scene flow is not the same as predicting motion vectors that minimize the distance between two point clouds. Each point in the first point cloud plus its flow vector may not necessarily correspond to a point in the second point cloud; rather it gives the corresponding location of the first point in the time frame of the second point cloud. However, given dense enough point clouds, predicting scene flow approximates minimizing the distance between the two point clouds. 

\medskip
\noindent \textbf{FlowNet3D Adapted (FN3DA)} \hspace{2pt}
We also take the FlowNet3D architecture and train it directly on our task. FlowNet3D only takes in two point clouds, so we train it on $x_{t-1}, x_t$ to predict $x_{t+1}$. This network falls under the category of PointNet++ with downsampling and can be seen as the two frame version of PN++ w/ DS. We use the same training procedure that we use for our models.

\subsection{Quantitative Results}

We evaluate our models on the nuScenes test set, consisting of about 70,000 frames preprocessed the same way as our training data. To measure the accuracy of our predictions, we use the CD and EMD between the predicted point clouds and the ground truth. Our results are shown in Figure \ref{fig:results} and Table \ref{table:performance}. FN3DOOB performs far worse than the other methods and skews the scale of Figure \ref{fig:results}, so we omit it to make the plot more interpretable. 

Among all the approaches, we find that FN3DA, PN++ w/ DS, and EC w/o DS achieve the lowest CD and EMD, with PN++ w/ DS performing marginally better. So for EdgeConv, not downsampling is important for strong performance, while for PointNet++, downsampling is actually more beneficial. We speculate this is due to the inherent difference in expressivity between the two architectures. The PointNet++ architecture benefits from downsampling because the memory efficiency allows it to utilize wider layers and learn more complex, hierarchical features. Indeed, with about the same memory consumption, our downsampling networks are 4 times wider than their non-downsampling counterparts. However, the EdgeConv architecture is already able to learn complex features on its own with smaller layers and therefore does not benefit as much from a larger network. On the other hand, the feature ambiguity caused by downsampling may be more harmful to EdgeConv, because resolving these ambiguities and upsampling in the feature space is more challenging than in Euclidean space. 

Although PN++ w/o DS and EC w/ DS perform better than the identity baseline in terms of CD, they have slightly higher EMD values. This is likely due to EMD's strong correlation with visual similarity. While our deep networks may learn point cloud dynamics, they often exhibit artifacts and struggle to replicate the clean appearance of a raw LiDAR scan. nuScenes high frame rate results in smaller motion between each frame. This allows the identity baseline to have exceptionally low EMD, as it maintains the clean appearance of a raw point cloud while not being penalized harshly for neglecting the dynamics of the scene. 

Lastly, we find that FN3DOOB is actually a destructive operation, increasing the CD and EMD more than the identity baseline. We believe this is due to the domain transfer from the synthetic FlyingThings3D dataset to real LiDAR scans, as well as the fundamental difference between the point cloud prediction and scene flow problem.

We also show the size, runtime, and memory usage of our models in Table \ref{table:performance}. Our models are implemented with PyTorch and tested on an Nvidia Titan RTX. The runtime and max memory allocated values are acquired with a batch size of 1, predicting $t+1$, on a point cloud with 22,000 points. The downsampling models tend to be significantly faster yet larger than their non downsampling counterparts. 

Based on this, we conclude that PN++ w/ DS and EC w/o DS are the most viable models. They both demonstrate high accuracy, but offer different computational advantages. PN++ w/ DS's runtime is about 2x faster, while EC w/o DS's model size is about 6x smaller. So if inference speed is more important, PN++ w/ DS should be used, but if model size is more important, then EC w/o DS should be used. 

While FN3DA performs competitively, PN++ w/ DS surpasses it in terms both speed and accuracy, so FN3DA offers no notable advantages. However, it achieves comparable accuracy to PN++ w/ DS while using only two frames, indicating that additional frames may only slightly improve performance. We speculate that conditioning on even more frames would improve accuracy by negligible amounts. 

\subsection{Visualizations}

We visualize some of our predictions for $t+5$ in Figure \ref{fig:visualization}. In this scene, the ego vehicle is driving past a truck parked next to a line of v-shaped columns. In the magnified portion of the visualization, the truck is in the top right, and the columns are along the left side. Among all the methods, PN++ w/ DS and EC w/o DS most accurately predict the dynamics of the truck and columns. The performance is corroborated in the error visualizations, which show green or yellow regions around the truck and columns for all the methods except for PN++ w/ DS and EC w/o DS. These methods instead show a nearly entirely purple point cloud, indicating close to 0 error. While we only visualize one scene here, we have done extensive qualitative testing on the entire nuScenes mini dataset (10 scenes) and have verified that on average, PN++ w/ DS and EC w/o DS outperform the other methods, corroborating our quantitative analysis. We include some of these visualizations in the supplementary section.

\subsection{Scene Flow Estimation}

We also show that our model is able to reasonably estimate scene flow. In Figure \ref{fig:flow_vis}, we visualize motion vectors predicted by our models and baselines for the same scene shown in Figure \ref{fig:visualization}, specifically highlighting the columns and the truck. Because the ego vehicle is moving forward, the columns and truck have motion vectors pointing backward. PN++ w/ DS and EC w/o DS perform the best, producing smooth, accurate scene flow, while the other methods either exhibit incoherent flow (FN3DOOB and PN++ w/o DS), or underestimate the magnitude (FN3DA, EC w/o DS), corroborating Figure \ref{fig:visualization}. More flow visualizations in diverse scenarios are available in the supplementary material. 

We would like to highlight that our self-supervised scene flow estimation is nontrivial. As mentioned before, minimizing the CD and EMD between two point clouds is not the same as minimizing scene flow. Directly learning self-supervised scene flow between time $t$ and $t+1$ using point cloud similarity metrics would result in degenerate outputs where the predicted vectors merely connect the points. In fact, there are a number of recent papers in the literature that try to regularize this ill-conditioned problem by adding smoothness constraints or cycle consistency \cite{pointpwcnet, justgo}. Our work addresses this by instead utilizing prior frames. Because the network is no longer given $t+1$, it cannot produce the degenerate solution. However, it is still given sufficient information on the scene’s dynamics in the prior frames to predict scene flow. Therefore, besides FN3DOOB, which is a trivial extension of FlowNet3D, the remaining approaches we describe are novel in the sense that they regularize the self-supervised scene flow problem in a new way. Here, we only qualitatively evaluate our estimated scene flow as a proof of concept; however, future work could build on this idea and produce more rigorous, quantitative analysis. 

\section{Conclusion}

In this paper we explore the task of point cloud prediction by designing a novel class of neural network architectures and training framework. We show that our top models (PN++ w/ DS and EC w/o DS) can generate convincing predictions of future point clouds, and that they are competitive with several strong baselines. Our visualizations help verify our findings and indicate that our models can be used to produce scene flow approximations. 

Our work has numerous applications and extensions, including self-supervised scene flow prediction, object tracking, temporal object detection, and vehicle control. Our use of CD and EMD loss functions may be improved by incorporating additional terms such as a perceptual loss or smoothness constraint. Lastly, our prediction accuracy can be further refined using RGB sensor fusion techniques. 

\medskip
\noindent \textbf{Acknowledgements}
The authors acknowledge the gracious support of Amazon for this project. 

{\small
\bibliographystyle{ieee}
\bibliography{ourbib}
}

\clearpage
\section{Supplementary}

\begin{table*}[h]
\begin{center}
\begin{tabularx}{\textwidth}{|>{\centering\arraybackslash}X||>{\raggedright\arraybackslash}X|>{\raggedright\arraybackslash}X|>{\raggedright\arraybackslash}X|>{\raggedright\arraybackslash}X|>{\raggedright\arraybackslash}X|}
\hline
\textbf{Module} & \textbf{PN++ w/ DS} & \textbf{PN++ w/o DS} & \textbf{EC w/ DS} & \textbf{EC w/o DS}\\
\hline
\hline
Feature Extraction 1 & \textbf{PointNet++}: r = 0.5, SR=0.25$\times$, mlp=[128,128] & 
    \textbf{PointNet++}: r=0.7, SR=1$\times$, mlp=[32,32] &
    \textbf{EdgeConv}: k=16, SR=0.25$\times$, mlp=[128,128] &
    \textbf{EdgeConv}: k=16, SR=1$\times$, mlp=[32,32]\\
\hline
Flow Embedding 1& \textbf{PointNet++}: r = 1.5, SR=1$\times$, mlp=[128] & 
    \textbf{PointNet++}: r=1, SR=1$\times$, mlp=[32] &
    \textbf{EdgeConv}: k = 16, SR=1$\times$, mlp=[128] &
    \textbf{EdgeConv}: k = 16, SR=1$\times$, mlp=[32]\\
\hline
Feature Extraction 2& \textbf{PointNet++}: r = 1, SR=0.25$\times$, mlp=[256,256] & 
    \textbf{PointNet++}: r = 0.7, SR=1$\times$, mlp=[64,64] &
    \textbf{EdgeConv}: k = 16, SR=0.25$\times$, mlp=[256,256] &
    \textbf{EdgeConv}: k = 16, SR=1$\times$, mlp=[64,64]\\
\hline
Flow Embedding 2& \textbf{PointNet++}: r = 3, SR=1$\times$, mlp=[256] & 
    \textbf{PointNet++}: r = 1, SR=1$\times$, mlp=[64] &
    \textbf{EdgeConv}: k = 16, SR=1$\times$, mlp=[256] &
    \textbf{EdgeConv}: k = 16, SR=1$\times$, mlp=[64]\\
\hline
Feature Extraction 3& \textbf{PointNet++}: r = 2, SR=0.2$\times$, mlp=[512] & 
    \textbf{PointNet++}: r = 0.7, SR=1$\times$, mlp=[128] &
    \textbf{EdgeConv}: k = 16, SR=0.2$\times$, mlp=[512] &
    \textbf{EdgeConv}: k = 16, SR=1$\times$, mlp=[128]\\
\hline
Refinement & \textbf{Upconv1}: k = 16, SR=5$\times$, SS=XYZ, mlp1=[512], mlp2=[512] &
    \textbf{mlp}: input = (feat1, feat2, feat3), widths = [512, 256, 128, 3] &
    \textbf{Upconv1}: k = 16, SR=5$\times$, SS=flow2, mlp1=[512], mlp2=[512]&
    \textbf{mlp}: input = (feat1, feat2, feat3), widths = [512, 256, 128, 3]\\
    & \textbf{Upconv2}: k = 16, SR=4$\times$, SS=XYZ, mlp1=[512], mlp2=[512] & &
    \textbf{Upconv2}: k = 16, SR=4$\times$, SS=flow1, mlp1=[512], mlp2=[512] & \\
    & \textbf{FeatProp}: SR=4$\times$, SS=XYZ, mlp=[256] & &
    \textbf{FeatProp}: SR=4$\times$, SS=XYZ, mlp=[256] & \\
    & \textbf{mlp}: [256, 128, 3] & &
    \textbf{mlp}: [256, 128, 3] & \\

\hline
\end{tabularx}
\end{center}
\caption{\textbf{Architecture details.} r = ball query radius, k = k for KNN grouping, SR = sampling rate, SS = sampling space, feat and flow refer to the output of the corresponding layer. }
\label{table:arch_details}
\end{table*}

Here we show our architecture details, the distribution of errors among each method, as well as additional visualizations of our models' prediction error and flow on different scenes and time steps. Both the distributions and the visualizations make it apparent that our models are able to outperform existing methods such as FlowNet3DOOB. 

\subsection{Architecture Details}
We describe the specific architecture details in Table \ref{table:arch_details}. The radii of the ball queries in the PointNet++ models were chosen to be commensurate to the receptive field of their EdgeConv counterparts (parameterized by k). As shown, we extend our architectures' choice of feature extractor to its flow embedding layers as well. 

\subsection{Error Distribution}
\begin{figure*}[b]
\begin{center}
\begin{subfigure}{\textwidth}
\includegraphics[width=\textwidth]{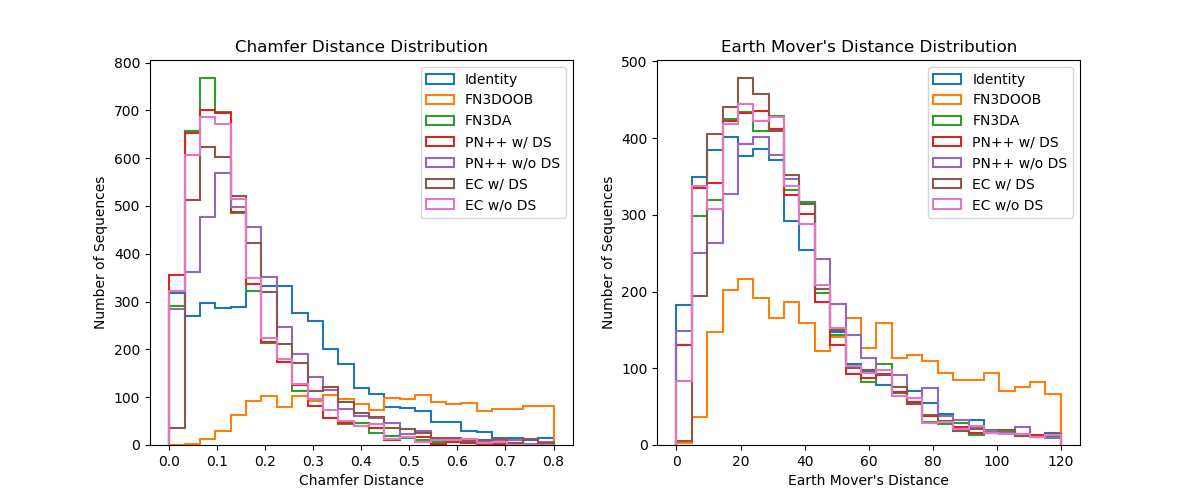}
\caption{Outliers removed}
\label{fig:hist_or}
\end{subfigure}
\begin{subfigure}{\textwidth}
\includegraphics[width=\textwidth]{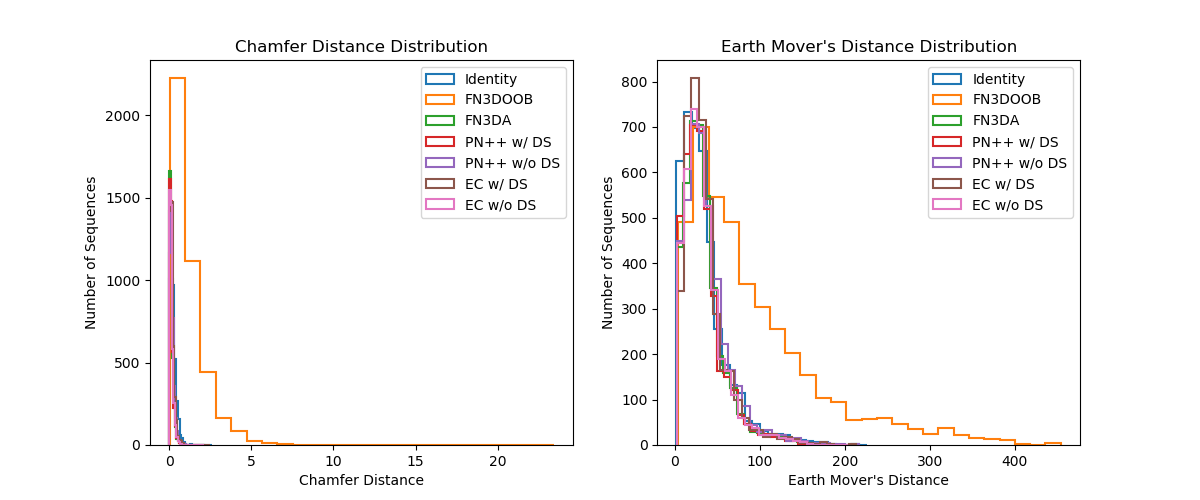}
\caption{With outliers}
\label{fig:hist_o}
\end{subfigure}
\end{center}
   \caption{\textbf{Distribution of Chamfer Distance and Earth Mover's Distance.} Histograms of errors for each method over 4000+ test samples. Each data point is the average of the error over a 5 frame sequence. The first two plots are have outliers removed and the other two show the entire distribution with the x-axis adjusted to range from 0 to the largest data point. }
\label{fig:hist}
\end{figure*}

In Figure \ref{fig:hist}, we can see that among all the methods, PN++ w/ DS and EC w/o DS have the highest concentration of points near 0 error and contain no prominent outliers. On the other hand, FN3DOOB performs far worse than the identity operation and contains some extreme outliers, especially in CD, although these points may not be discernible on the plot due to the scaling of the y-axis. 

\subsection{Error Visualization}
\begin{figure*}[hbtp]
\begin{center}
    \includegraphics[width=\textwidth, center]{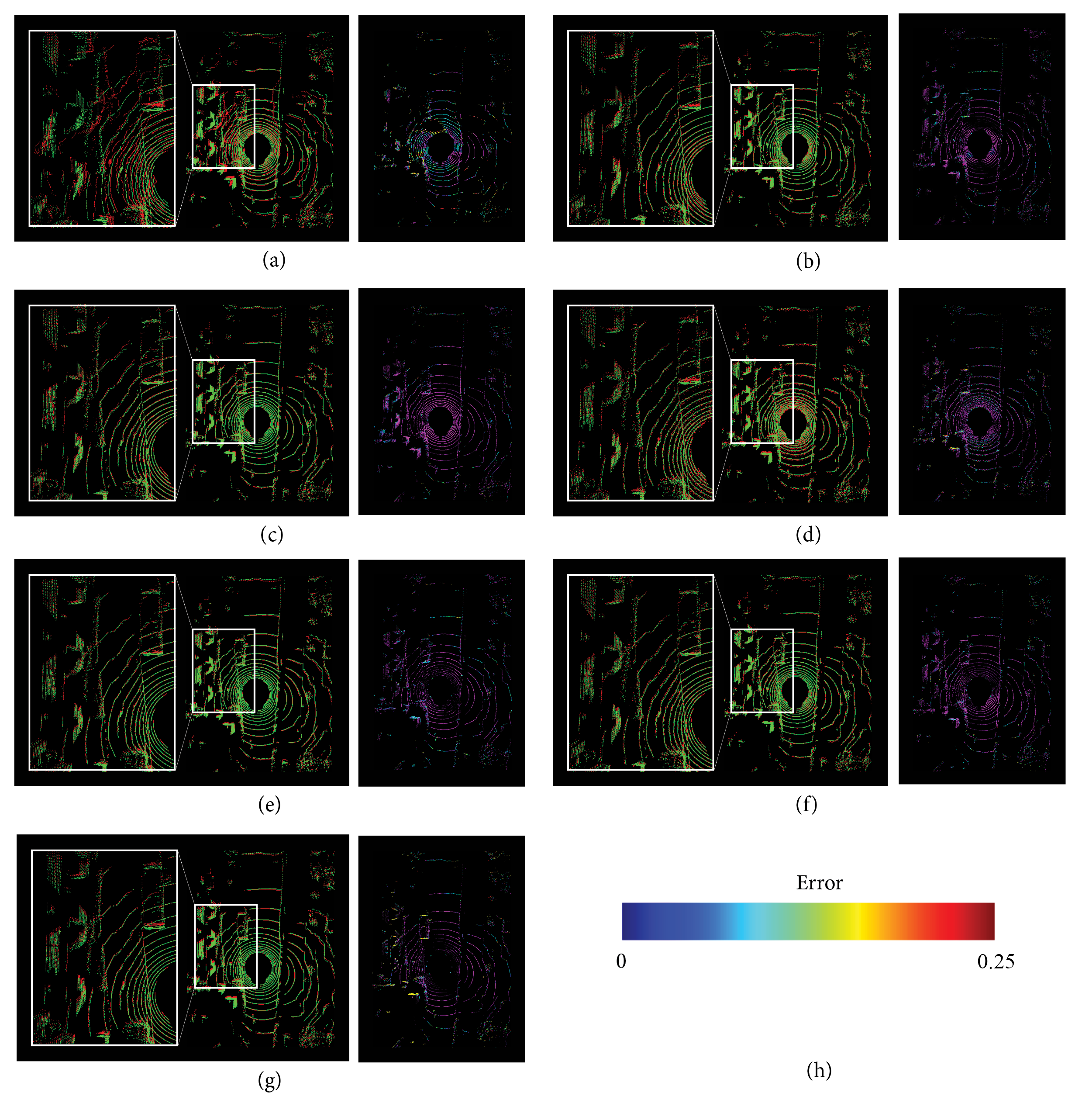}
\end{center}
\caption{Visualization of predictions for $t+1$ on the same scene shown in Figure \ref{fig:visualization}. (a) FN3DOOB, (b) FN3DA, (c) PN++ w/ DS, (d) PN++ w/o DS, (e) EC w/ DS, (f) EC w/o DS, (g) Identity, (h) scale of error as described in Figure \ref{fig:visualization}. }
\label{fig:supp_vis_1}
\end{figure*}

\begin{figure*}[hbtp]
\begin{center}
    \includegraphics[width=\textwidth, center]{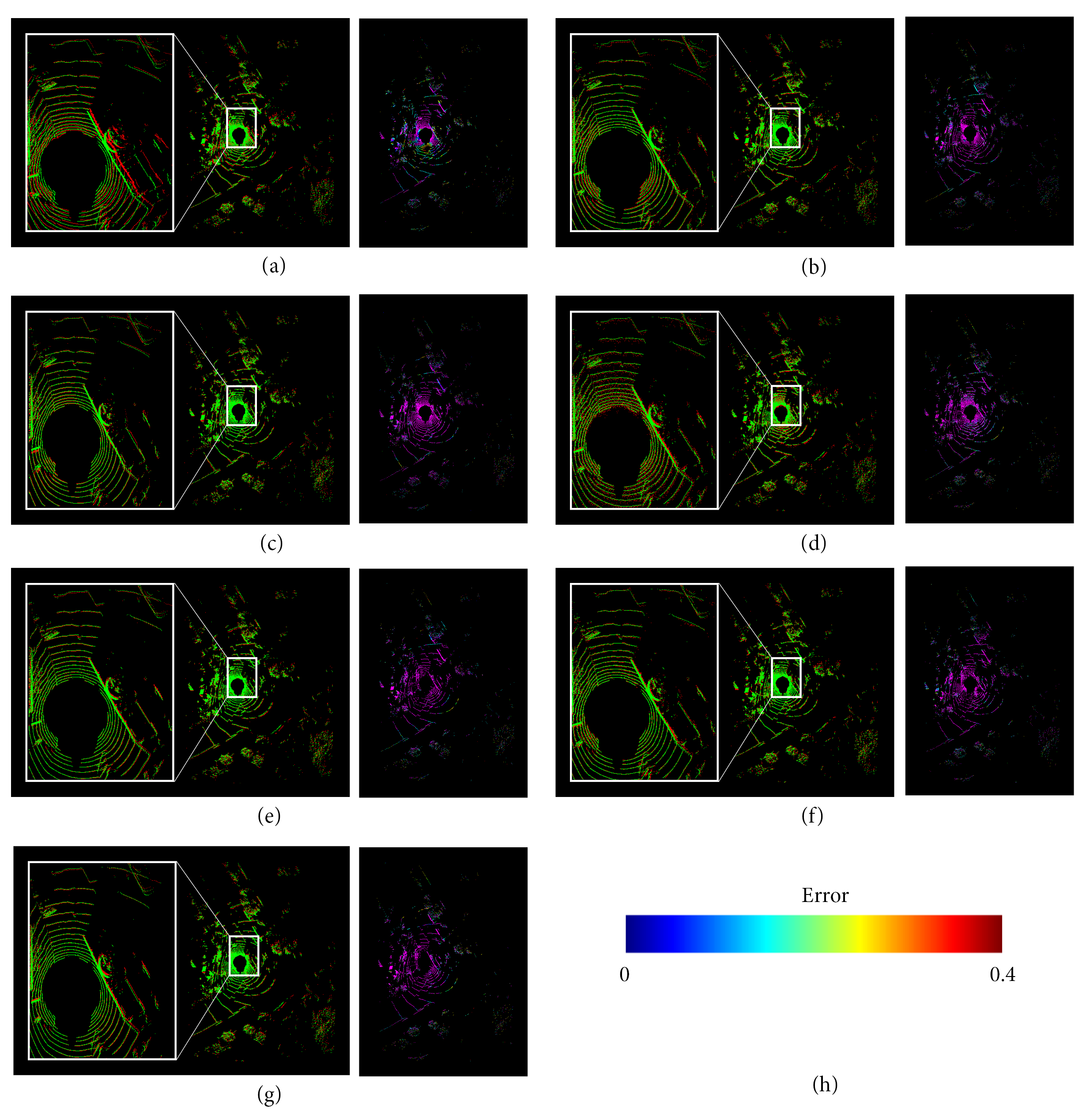}
\end{center}
    \caption{Visualization of predictions for $t+1$ on an additional scene. (a) FN3DOOB, (b) FN3DA, (c) PN++ w/ DS, (d) PN++ w/o DS, (e) EC w/ DS, (f) EC w/o DS, (g) Identity, (h) scale of error as described in Figure \ref{fig:visualization}.}
\label{fig:supp_vis_2}
\end{figure*}

\begin{figure*}[hbtp]
\begin{center}
    \includegraphics[width=\textwidth, center]{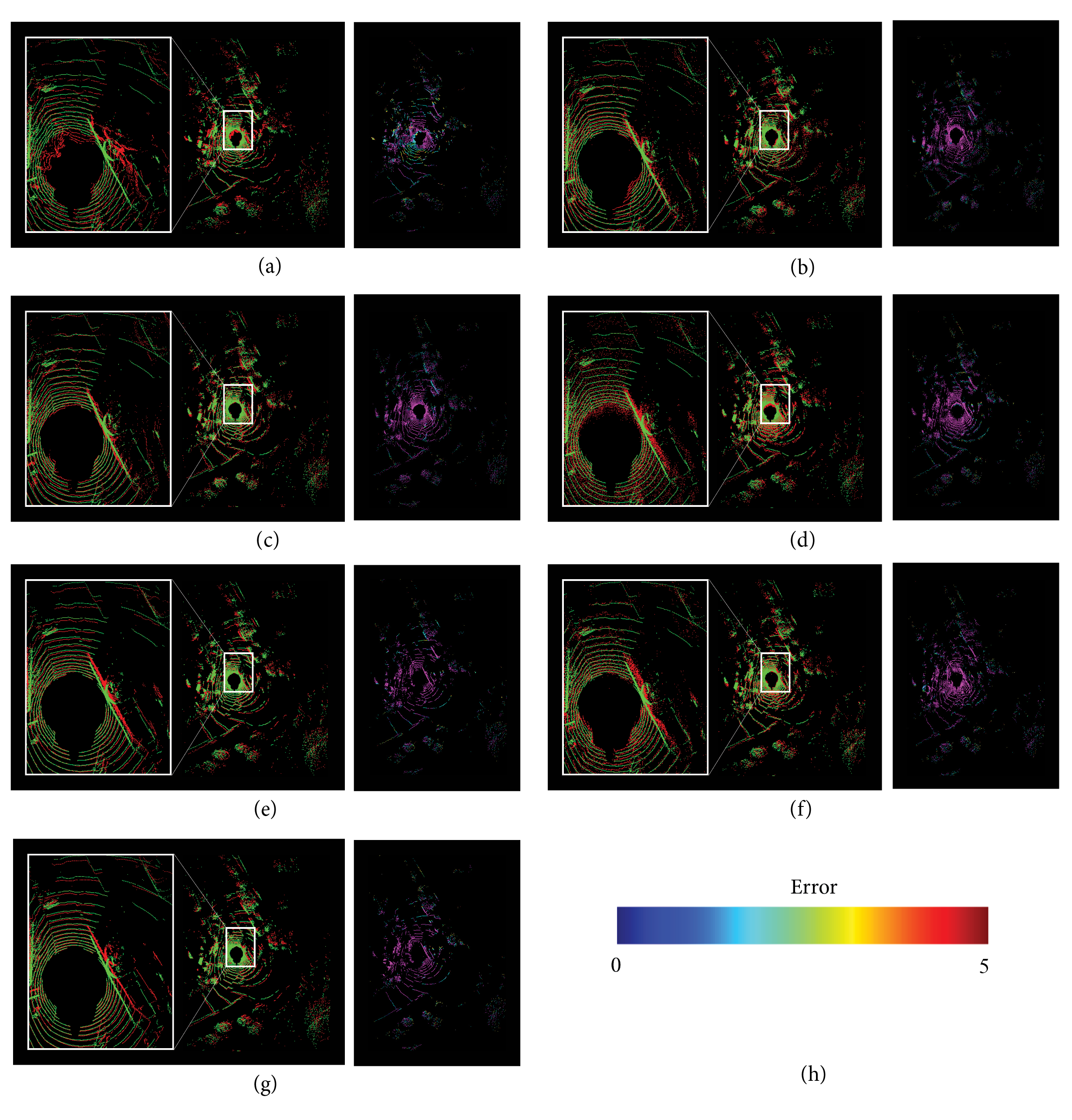}
\end{center}
    \caption{Visualization of predictions for $t+5$ on an additional scene. (a) FN3DOOB, (b) FN3DA, (c) PN++ w/ DS, (d) PN++ w/o DS, (e) EC w/ DS, (f) EC w/o DS, (g) Identity, (h) scale of error as in Figure \ref{fig:visualization}.}
\label{fig:supp_vis_3}
\end{figure*}

In Figure \ref{fig:supp_vis_1}, we show the same scene from Figure \ref{fig:visualization}, but predicted for $t+1$, and in Figures \ref{fig:supp_vis_2} and \ref{fig:supp_vis_3}, we show $t+1$ and $t+5$ on another scene. In Figures \ref{fig:supp_vis_2} and \ref{fig:supp_vis_3}, the ego vehicle is making a left turn at an intersection. At the top left of the the zoomed in view there is a van driving in front of the vehicle, while at the bottom left pedestrians cross the street behind it. This complex movement of the ego coordinate frame combined with the irregular geometries present make this second scene more challenging. PN++ w/ DS best predicts the position of the large barrier to the right of the vehicle, and while all of our methods seem to make reasonable predictions of the van and pedestrians, EC w/ DS seems to do it the most accurately. Interestingly, we noticed that EC w/ DS tends to preserve the orderly appearance of a raw LiDAR scan despite generally having higher EMD. 

\subsection{Scene Flow Visualization}
\begin{figure*}[t]
\begin{center}
    \includegraphics[width=\textwidth, center]{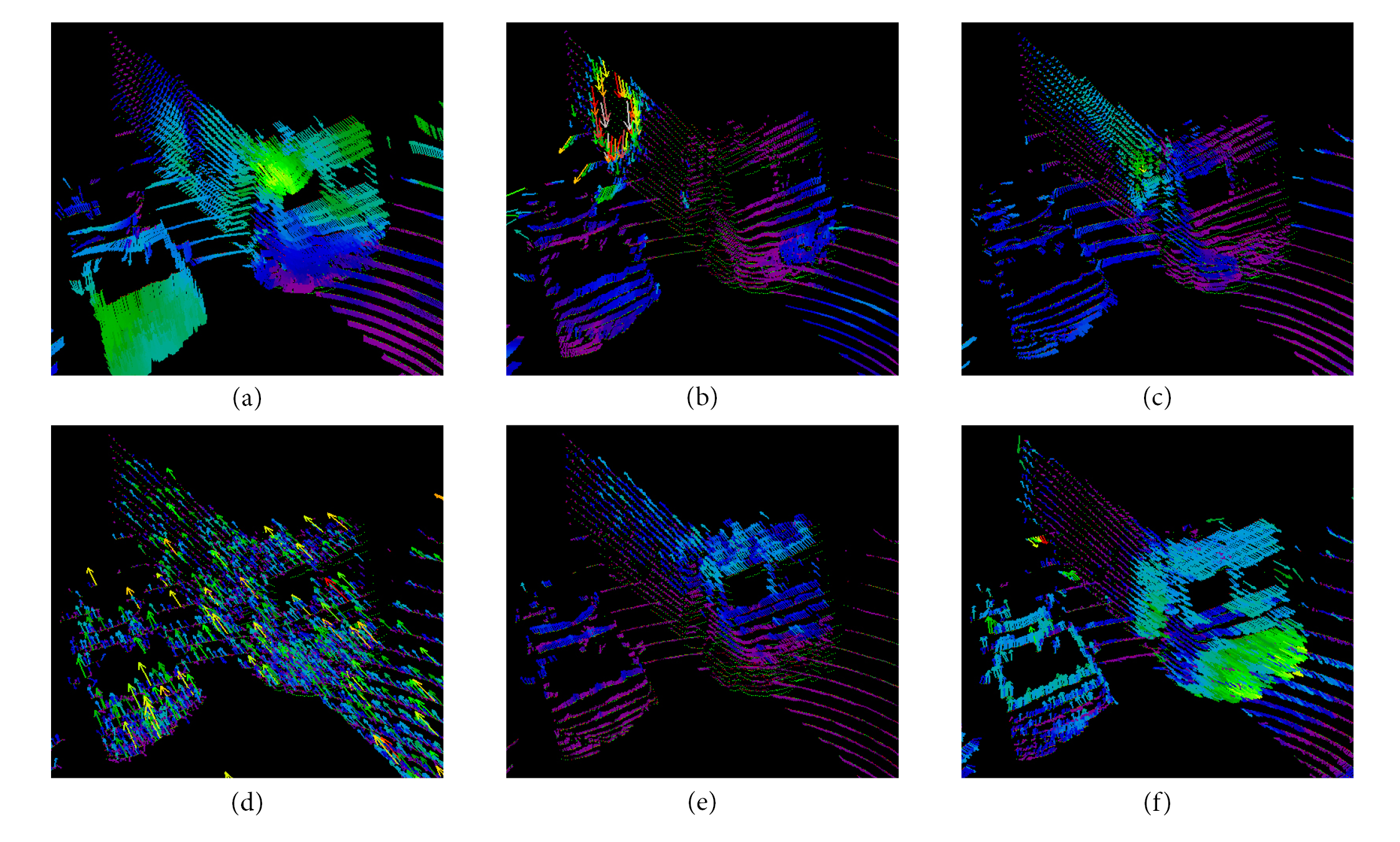}
    \vspace{-30pt}
\end{center}
    \caption{Additional flow visualization. (a) FN3DOOB, (b) FN3DA, (c) PN++ w/ DS, (d) PN++ w/o DS, (e) EC w/ DS, (f) EC w/o DS. The vectors are colored on the same scale as the visualization in the main paper. In this scene, the ego vehicle is stationary while a truck approaches from behind. The only moving object is the truck, so the true flow vectors would indicate the truck moving forward while all other points have no motion. EC w/ DS and FN3DOOB most accurately capture the motion of the truck. However, FN3DOOB overestimates the motion of the car next to the truck and exhibits flow discontinuities, shown by the yellow patch at the top left corner of the truck and the decreasing magnitude near the ground.}
\label{fig:supp_flow_vis_1}
\end{figure*}

\begin{figure*}[b]
\begin{center}
    \includegraphics[width=\textwidth, center]{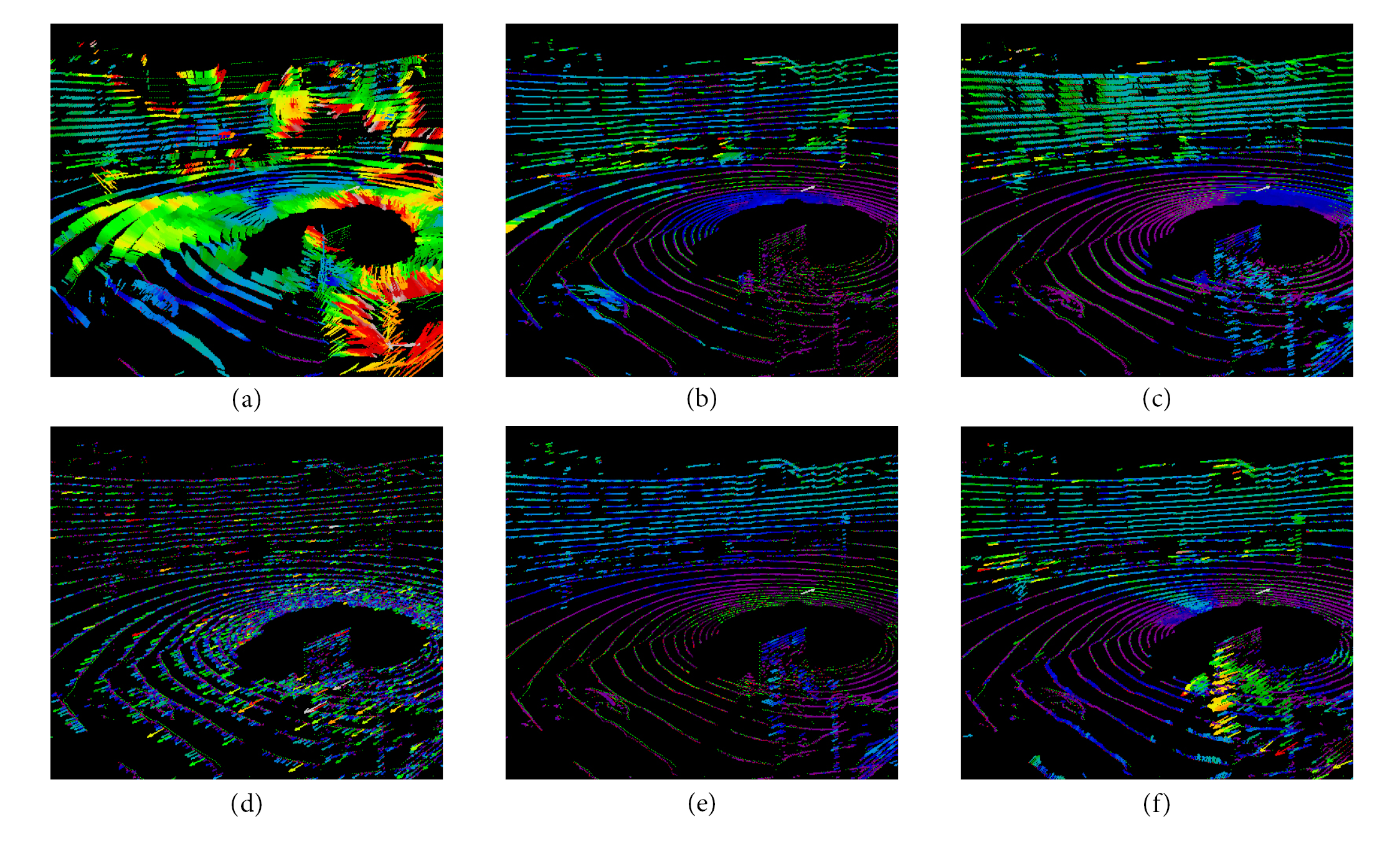}
    \vspace{-30pt}
\end{center}
    \caption{Additional flow visualization. (a) FN3DOOB, (b) FN3DA, (c) PN++ w/ DS, (d) PN++ w/o DS, (e) EC w/ DS, (f) EC w/o DS. The vectors are colored on the same scale as the visualization in the main paper. In this scene, the ego vehicle is driving forward while a car, located in the bottom left of the image, follows it from behind at the same speed. So the background should have moderate backward motion vectors, while the trailing car should show little movement. PN++ w/ DS, EC w/ DS, and EC w/o DS are able to reasonably capture this. FN3DA fails to capture the small relative motion of the trailing car.}
\label{fig:supp_flow_vis_2}
\end{figure*}

\begin{figure*}[t]
\begin{center}
    \includegraphics[width=\textwidth, center]{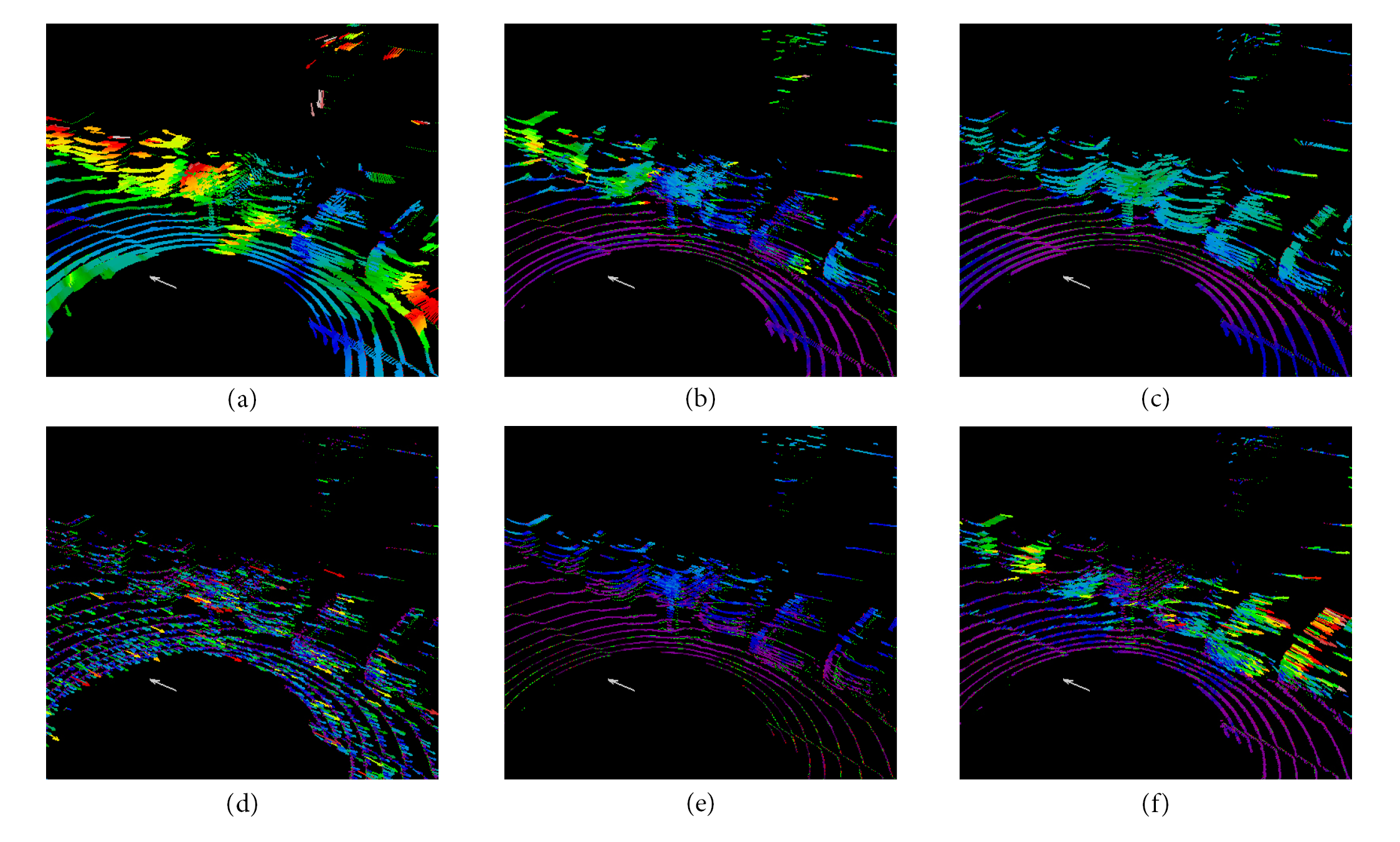}
    \vspace{-25pt}
\end{center}
    \caption{Additional flow visualization. (a) FN3DOOB, (b) FN3DA, (c) PN++ w/ DS, (d) PN++ w/o DS, (e) EC w/ DS, (f) EC w/o DS. The vectors are colored on the same scale as the visualization in the main paper. In this scene, the ego vehicle is driving past a full parking lot, shown in the visualization. The true flow vectors would show the cars moving backward relative to the ego vehicle. Here, PN++ w/ DS performs the best, producing smooth, accurate motion vectors for the parked cars. }
\label{fig:supp_flow_vis_3}
\end{figure*}

\begin{figure*}[b]
\begin{center}
    \includegraphics[width=\textwidth, center]{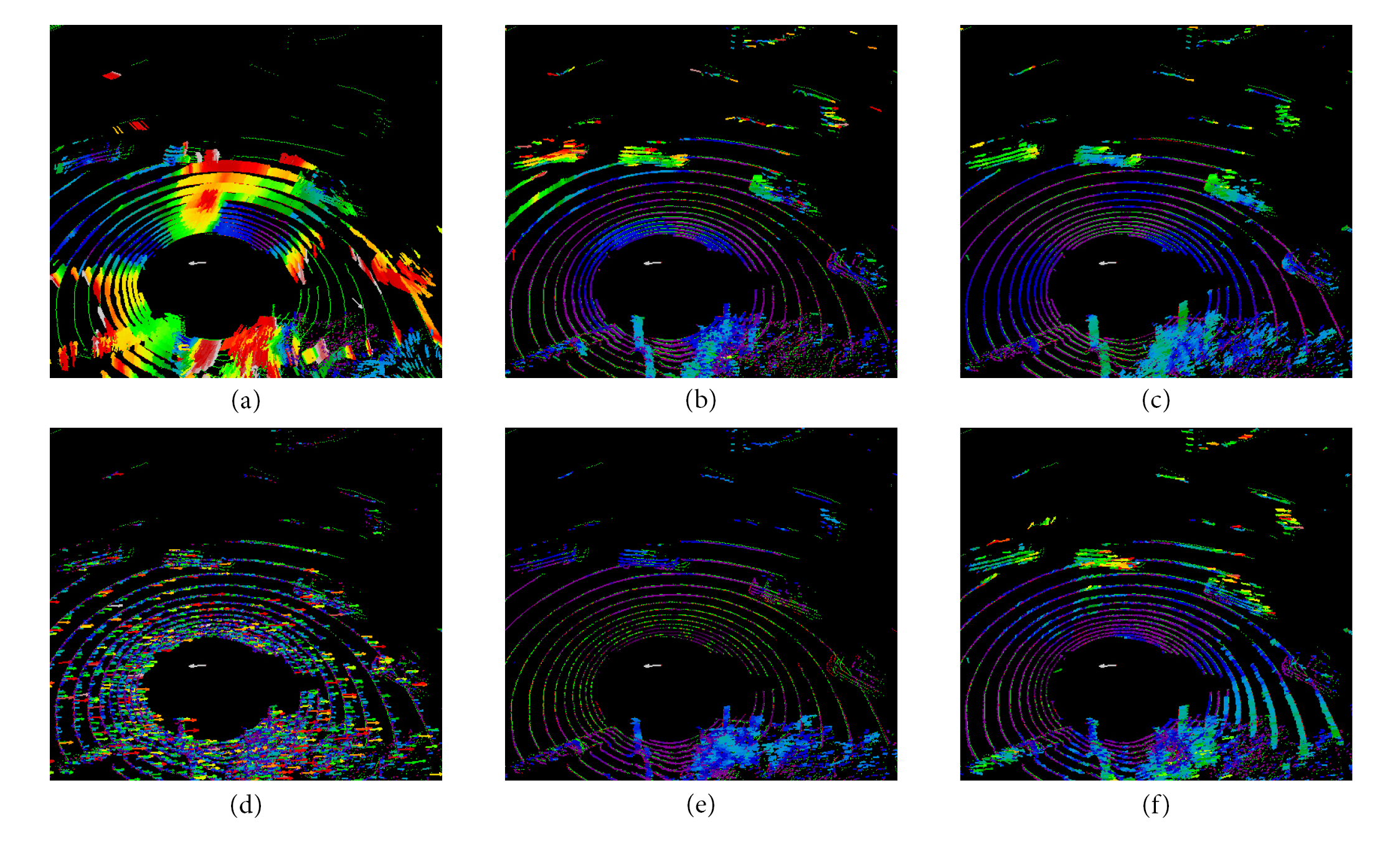}
    \vspace{-25pt}
\end{center}
    \caption{Additional flow visualization. (a) FN3DOOB, (b) FN3DA, (c) PN++ w/ DS, (d) PN++ w/o DS, (e) EC w/ DS, (f) EC w/o DS. The vectors are colored on the same scale as the visualization in the main paper. In this scene, the ego vehicle is making a left turn while cars coming from its left turn right, making the same turn but in the opposite direction. Three of these cars are shown moving from left to right across the visualizations. Note this scene is in Singapore, where vehicles drive on the left side of the road. At the same time, some cars are stopped behind the ego vehicle, one of which is shown near the bottom right of the visualization. FN3DA, PN++ w/ DS, and EC w/o DS are all able to predict the motion of the right turning cars, while producing small motion vectors for the stopped cars.}
\label{fig:supp_flow_vis_4}
\end{figure*}

In Figures \ref{fig:supp_flow_vis_1}, \ref{fig:supp_flow_vis_2}, \ref{fig:supp_flow_vis_3}, and \ref{fig:supp_flow_vis_4} we show some additional flow visualizations. Specific comments on the visualizations can be found in the captions. These scenes were taken from the nuScenes Mini Dataset and show that on average, PN++ w/ DS produces the most coherent predictions, with EC w/o DS also performing well, corroborating our quantitative analysis. 

\subsection{Code}
Our code will be released at https://github.com/davezdeng8/tlfpad. EC w/o DS is implemented based on the official PyTorch implementation of DGCNN, and the other models are implemented based on a PyTorch implementation of FlowNet3D. While one would expect EC w/ DS to use less memory than EC w/o DS, due to the implementation difference, EC w/o DS uses less memory as indicated in Table \ref{table:performance}. 

\end{document}